\documentclass[sn-basic,iicol]{sn-jnl}

\usepackage{colortbl,makecell}
\usepackage{multirow,array}
\usepackage{amsmath}

\makeatletter
\newcommand\footnoteref[1]{\protected@xdef\@thefnmark{\ref{#1}}\@footnotemark}

\jyear{2022}%

\begin{document}

\title[Controlled Gaussian Process Dynamical Models with Application to Robotic Cloth Manipulation]{Controlled Gaussian Process Dynamical Models with Application to Robotic Cloth Manipulation}

\author*[1]{\fnm{Fabio} \sur{Amadio}}\email{fabio.amadio@phd.unipd.it}

\author[2]{\fnm{Juan Antonio} \sur{Delgado-Guerrero}}\email{jdelgado@iri.upc.edu}

\author*[2]{\fnm{Adrià} \sur{Colomé}}\email{acolome@iri.upc.edu}

\author[2]{\fnm{Carme} \sur{Torras}}\email{torras@iri.upc.edu}

\affil*[1]{\orgdiv{Department of Information Engineering}, \orgname{Università degli Studi di Padova}, \orgaddress{\street{Via Gradenigo 6/B}, \city{Padova}, \postcode{35131}, \country{Italy}}}

\affil[2]{\orgdiv{Institut de Robòtica i Informàtica Industrial}, \orgname{CSIC-UPC}, \orgaddress{\street{C/ Llorens i Artigas 4-6}, \city{Barcelona}, \postcode{08028},  \country{Spain}}}

\abstract{
Over the last years, significant advances have been made in robotic manipulation, but still, the handling of non-rigid objects, such as cloth garments, is an open problem.
Physical interaction with non-rigid objects is uncertain and complex to model. Thus, extracting useful information from sample data can considerably improve modeling performance. However, the training of such models is a challenging task due to the high-dimensionality of the state representation.
In this paper, we propose Controlled Gaussian Process Dynamical Model (CGPDM) for learning high-dimensional, nonlinear dynamics by embedding it in a low-dimensional manifold.
A CGPDM is constituted by a low-dimensional latent space, with an associated dynamics where external control variables can act and a mapping to the observation space.
The parameters of both maps are marginalized out by considering Gaussian Process (GP) priors. Hence, a CGPDM projects a high-dimensional state space into a smaller dimension latent space, in which it is feasible to learn the system dynamics from training data.
The modeling capacity of CGPDM has been tested in both a simulated and a real scenario, where it proved to be capable of generalizing over a wide range of movements and confidently predicting the cloth motions obtained by previously unseen sequences of control actions.
}


\maketitle

\bmhead{Funding}
This work was partially developed in the context of the project CLOTHILDE ("CLOTH manIpulation Learning from DEmonstrations"), which has received funding from ERC under the European Union’s Horizon 2020 research and innovation program (Advanced Grant agreement No 741930).

\bmhead{Competing Interests}
The authors declare that they have no competing interests.

\bmhead{Author Contributions}
Fabio Amadio, Juan Antonio Delgado-Guerrero and Adrià Colomé conceived the presented idea.
Fabio Amadio developed the theory, implemented the code and carried out the numerical experiments.
Fabio Amadio took the lead in writing the manuscript.
Carme Torras supervised the project.
All authors provided critical feedback and helped shape the research, analysis and manuscript.

\bmhead{Acknowledgments}
We would like to thank Adrià Luque Acera for his help with the data collection in the real-world experiment, and Ce Xu for his useful feedback during code development.

\section{Introduction}\label{sec1}
Robotic cloth manipulation has a wide range of applications, from textile industry to assistive robotics \cite{laundryFolding2011Bersch, towelFolding2012Abbeel, lakshmanan2013constraint, DeformableSurvey, BenchmarkBimanual, GraspingCentered}. However, the complexity of cloth behaviour results in a high uncertainty in the state transition given a certain action. This uncertainty is what makes manipulating cloth much more challenging than handling rigid objects. Intuitively, learning the cloth's dynamics is the solution to reduce such uncertainty. In literature, we can find several cloth models that simulate the internal cloth state \cite{Cloth1,Cloth3,Cloth2}. They represent cloth as a mesh of material points, and simulate their behaviour taking into account physical constraints. However, fitting those models to real data can be a complex task. Moreover, such models need not only to behave similarly enough to the cloth garment, but to have a tractable dimensionality, for computational reasons. As an example, an $8\times 8$ mesh representing a square towel results in a $192$-dimensional manifold. Such dimensionality is unmanageable, not only in terms of computational costs, but also for building a tractable state-action space policy.  Such is the case of \cite{MPCCloth}, where simulated results are obtained after hours of computations.

Hence, Dimensionality Reduction (DR) methods can be very beneficial.
In \cite{ColomeTRO}, linear DR techniques were used for learning cloth manipulation by biasing the latent space projection with each execution's performance. Nonlinear methods, such as Gaussian Process Latent Variable Models (GPLVMs) \cite{lawrence2005probabilistic} have also been applied for this purpose. In \cite{Koganti}, GPLVM was employed to project task-specific motor-skills of the robot onto a much smaller state representation, whereas in \cite{juananCovariate} a GPLVM was also used to represent a robot manipulation policy in a latent space, taking contextual features into account. However, these approaches focus the dimensionality reduction on the robot action characterization, rather than on the manipulated object's dynamics.
Instead, in \cite{koganti2017bayesian} a GPLVM learns a latent representation of the cloth state from point clouds. However, such approach did not consider the cloth handling task dynamics, limiting the application to quasi-static manipulations.

In this paper, we assume to have recorded data from several cloth motions, as a time-varying mesh of points.
To fit such data into a tractable dynamical model, we consider Gaussian Process Dynamical Models (GPDMs), first introduced in \cite{wang2005gaussian}, which are an extension of the GPLVM structure explicitly oriented to the analysis of high-dimensional time series. GPDMs have been applied in several different fields, from human motion tracking \cite{wang2007gaussian, urtasun20063d} to dynamic texture modeling \cite{zhu2016dynamic}. In the context of cloth manipulation, GPDMs were adopted in \cite{koganti2015cloth} to learn a latent model of the dynamics of a cloth handling task. However, this framework, as it stands, lacks in its structure a fundamental component to correctly describe the dynamics of a system, namely control actions, limiting generalization capacity.
\begin{figure}[t]
\centering
\includegraphics[width=\linewidth]{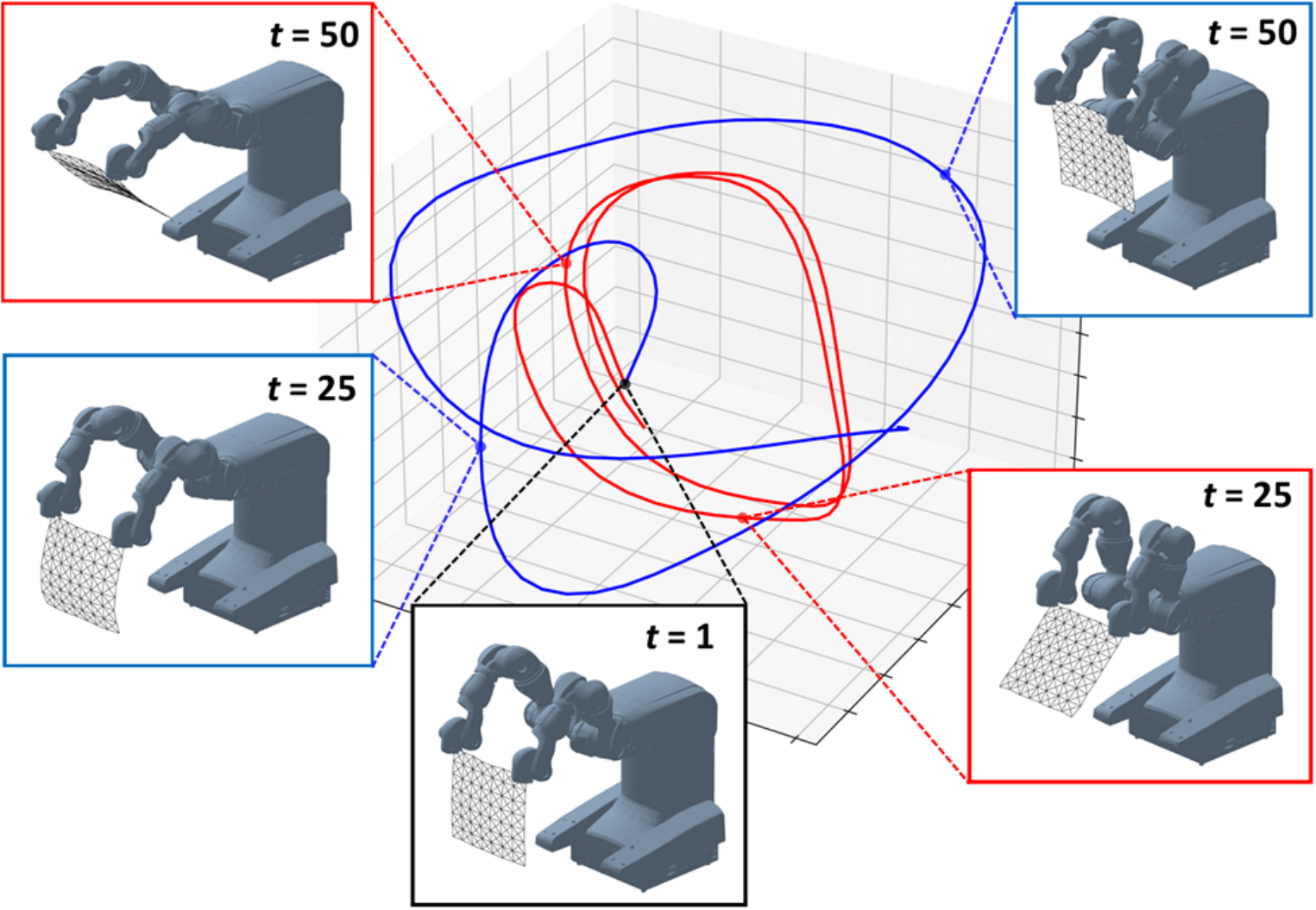}
\caption{Latent trajectories predicted by a trained CGPDM in response to two different sequences of unseen actions. Each latent state has associated a particular configuration of the cloth model (some of them are shown as an example)}
\label{fig:latent mapping}
\end{figure}

Therefore, we propose here an extension of the GPDM structure, that takes into account the influence of external control actions on the modeled dynamics. We call it \textit{Controlled Gaussian Process Dynamical Model} (CGPDM). In this new version, control actions directly affect the dynamics in the latent space. Thus, a CGPDM, trained on a sufficiently diverse set of interactions, is able to  predict the effects of control actions never experienced before inside a space of reduced dimensions, and then reconstruct high-dimensional motions by projecting the latent state trajectories into the observation space.
CGPDM has proved capable of fitting different types of cloth movements, in both a simulated and a real cloth manipulation scenario, and being able to predict the results of control actions never seen during training (example reported in Fig. \ref{fig:latent mapping}).
Finally, we compared two possible CGPDM parameterizations.
The first is a straightforward extension of standard GPDM, whereas in the second we propose to employ squared exponential (SE) kernels with automatic relevance determination (ARD) \cite{neal2012bayesian} and inhomogeneous linear kernels, together with tunable dynamical map scaling factors, obtaining a better accuracy and generalization, especially in the low-data regime.

To summarize, the main contributions of this article are:
\begin{itemize}
    \item The proposal of the CGPDM structure, an extension of the GPDM capable of taking into account the presence of exogenous inputs.
    \item The definition of a more rich parameterization able to achieve better accuracy and generalization w.r.t. the standard structure previously employed in the GPDM context.
    \item The successful application of the proposed CGPDM to (both simulated and real) dynamic robotic cloth manipulation problems.
\end{itemize}

The remainder of the paper is structured as follows. Sec.~\ref{sec:Methods} provides the details of the proposed CGPDM approach.
Results obtained by CGPDM in cloth dynamics modeling are described in Sec.~\ref{sec:results}, both in simulation and in a real case scenario.
Finally, the obtained results are discussed in Sec.~\ref{sec:discussion} and conclusions are drawn in Sec.~\ref{sec:conclusion}.

\section{Methods}\label{sec:Methods}
This section thoroughly describes the proposed method.
We start by providing some background notions about the models we build on top: GP, GPLVM, and GPDM (Subsec.~\ref{subsec:Pre}).
Then, we present the CGPDM (Subsec.~\ref{subsec:CGPDM}), detailing the structure of its latent and dynamics maps.
In particular, we present two alternative CGPDM structures: \textit{naive} and \textit{advanced}.
The first is a straightforward inclusion of exogenous inputs into standard GPDM, while the latter is the proposed CGPDM characterized by a richer parameterization.
Finally, we conclude by describing the model training and prediction procedures (Subsec.~\ref{subsec:CGPDM learning and pred}).

\subsection{Background: From GP to GPDM}\label{subsec:Pre}
GPs \cite{williams2006gaussian} are the infinite-dimensional generalization of multivariate Gaussian distributions. They are defined as infinite-dimension stochastic processes such that, for any finite set of input locations $\mathbf{x}_1, ..., \mathbf{x}_n$, the random variables $f(\mathbf{x}_1), ..., f(\mathbf{x}_n)$ have joint Gaussian distributions. A GP is defined by its mean function $m(\mathbf{x})$ and kernel $k(\mathbf{x}, \mathbf{x}')$, that must be a symmetric and positive semi-definite function. Usually GPs are denoted as $f(\mathbf{x}) \sim \mathcal{GP}(m(\mathbf{x}), k(\mathbf{x}, \mathbf{x}'))$.

GPs can be used for regression models of the form $y = f(\mathbf{x}) + \varepsilon$, with $\varepsilon$ an i.i.d. Gaussian noise, as they provide closed formulae to predict new target $y^*$, given new input $\mathbf{x}^*$. GP regression has been widely applied as a data-driven tool for dynamical system identification \cite{kocijan2005dynamic}, usually describing each state by its own GP. Nevertheless, such approach struggles to scale to high-dimensional systems. Thus, DR strategies must be considered.

GPLVMs \cite{lawrence2005probabilistic, GPLVMsurvey} emerged as feature extraction methods that can be used as multiple-output GP regression models. These models, under a DR perspective, associate and learn low-dimensional representations of higher-dimensional observed data, assuming that observed variables are determined by the latent ones. Finally, GPLVMs provide, as a result of an optimization, a mapping from the latent space to the observation space, together with a set of latent variables representing the observed values. However, GPLVMs are not explicitly thought to deal with time series, where a dynamics relate the values observed at consecutive time steps.

Thus, \cite{wang2005gaussian} first introduced Gaussian Process Dynamical Models (GPDM), an extension of the GPLVM structure explicitly oriented to the analysis of high-dimensional time series. A GPDM entails essentially two stages: (i) a latent mapping that projects high-dimensional observations to a low-dimensional latent space; (ii) a discrete-time Markovian dynamics that captures the evolution of the time series inside the reduced latent space.
GPs are used to model both maps.

\subsection{Controlled GPDM}\label{subsec:CGPDM}
Let us consider a system governed by an unknown dynamics. At each time step $t$, $\boldsymbol{u}_t \in \mathbb{R}^E$ represents the applied control action and $\boldsymbol{y}_t \in \mathbb{R}^D$ the observation.
For high-dimensional observation spaces, it could be unfeasible to directly model the evolution of a sequence of observations in response to a series of inputs.
For instance, in the case of a robot moving a piece of cloth, we can consider as control actions $\boldsymbol{u}_t$ the instantaneous movement of the end-effector, while the observations $\boldsymbol{y}_t$ could be the coordinates of a mesh of material points, representing the cloth configuration.
In this context, it could be convenient to capture the dynamics of the system in a low-dimensional latent space $\mathbb{R}^d$, with $d<<D$. Let $\boldsymbol{x}_t \in \mathbb{R}^d$ be the latent state associated with $\boldsymbol{y}_t$.
We propose to use a variation of the GPDM that keeps into account the influence of control actions, while maintaining the dimensionality reduction properties of the original model. We call it \textit{Controlled Gaussian Process Dynamical Model} (CGPDM).

\begin{figure}[t]
\centering
\includegraphics[width=\linewidth]{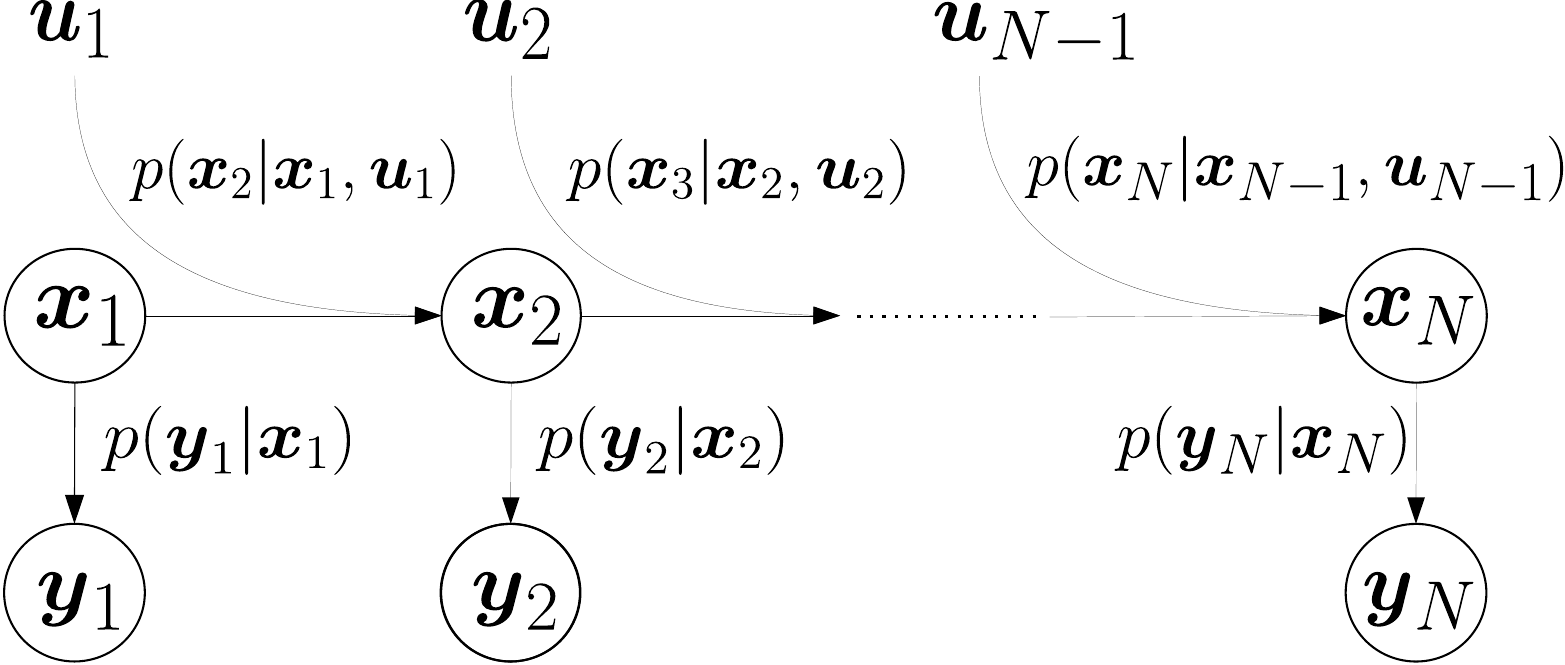}
\caption{
Symbolic representation of a CGPDM rollout along $N$ time steps. Note how output $\boldsymbol{y}$ depends exclusively on the latent state $\boldsymbol{x}$, while control action $\boldsymbol{u}$ influences only the latent dynamics
}
\label{fig:mdp_scheme}
\end{figure}

A CGPDM consists of a latent map \eqref{eq: latent mapping} projecting observations $\boldsymbol{y}_t$ into latent states $\boldsymbol{x}_t$, and a dynamics map \eqref{eq: markov dynamics} that describes the evolution of $\boldsymbol{x}_t$, subject to $\boldsymbol{u}_t$.
We denote the two maps as,
\begin{equation}\label{eq: latent mapping}
    \boldsymbol{y}_t = g(\boldsymbol{x}_t) + \boldsymbol{n}_{y,t}\text{,}
\end{equation}
\begin{equation}\label{eq: markov dynamics}
    \boldsymbol{x}_{t+1} - \boldsymbol{x}_t = h(\boldsymbol{x}_t, \boldsymbol{u}_t) + \boldsymbol{n}_{x,t}\text{.}
\end{equation}
where $\boldsymbol{n}_{y,t}$ and $\boldsymbol{n}_{x,t}$ are two zero-mean isotropic Gaussian noise processes, while $g$ and $h$ are two unknown functions. 
Differently from original GPDM, here the latent transition function \eqref{eq: markov dynamics} is also influenced by exogenous control inputs $\mathbf{u}_t$.
Note that we consider $\boldsymbol{x}_{t+1} - \boldsymbol{x}_t$ to be the output of the CGPDM dynamic map, \cite{wang2007gaussian} suggested that this choice can improve latent trajectories smoothness.
In the following, we report how we modeled \eqref{eq: latent mapping} and \eqref{eq: markov dynamics} by means of GPs, while Fig. \ref{fig:mdp_scheme} illustrates the relation assumed by CGPDM between the latent, input, and output spaces along $N$ time steps.

\subsubsection{Latent variable mapping}\label{subsubsec:latent map}
Each component of the observation vector $\boldsymbol{y}_t = [y_t^{(1)}, \dots, y_t^{(D)}]^T$ can be modeled a priori as a zero-mean GP that takes as input $\boldsymbol{x}_t$, for $t=1,\dots,N$.
Let $\mathbf{Y} = [ \boldsymbol{y}_1,\dots, \boldsymbol{y}_N]^T \in \mathbb{R}^{N \times D}$ be the matrix that collects the set of $N$ observations, and $\mathbf{X} = [ \boldsymbol{x}_1,\dots, \boldsymbol{x}_N]^T \in \mathbb{R}^{N \times d}$ be the matrix of associated latent states. We denote with $\mathbf{Y}_{:,j}$ the vector containing the $j$-th components of all the $N$ observations.
Then, if we assume that the $D$ observation components are independent variables, the probability over the whole set of observations can be expressed by the product of the $D$ GPs.
In addition, if we choose the same kernel function $k_y(\cdot,\cdot)$ for each GP, differentiated only through a variable scaling factor $w_{y,j}^{-2}$, with $j=1,\dots,D$, the joint likelihood over the whole set of observations is given by
\begin{multline}\label{eq: latent simplified density}
    p(\mathbf{Y}\vert  \mathbf{X}) = \frac{\vert \mathbf{W}_y\vert ^N}{\sqrt{(2\pi)^{ND} \vert \mathbf{K}_y(\mathbf{X})\vert ^D}} \cdot\\
     \text{exp}\left( -\frac{1}{2} \text{tr}\left(\left(\mathbf{K}_y(\mathbf{X})\right)^{-1} \mathbf{Y} \mathbf{W}_y^2 \mathbf{Y}^T\right)\right)\text{,}
\end{multline}
where $\mathbf{W}_y=\text{diag}(w_{y,1},\dots, w_{y,D})$, $\mathbf{K}_y(X)$ is the covariance matrix defined element-wise by $k_y(\cdot,\cdot)$. Independence assumption may be relaxed by applying coregionalization models \cite{vectorKernels}, at the cost of greater computational demands.
In previous GPDM works \cite{wang2005gaussian, wang2007gaussian, urtasun20063d}, the GPs of the latent map were equipped with an isotrophic SE kernel,
\begin{multline}\label{eq: old y kernel}
k_y'(\boldsymbol{x}_r, \boldsymbol{x}_s) = \text{exp}\left(-\frac{\beta_1}{2}\vert \vert \boldsymbol{x}_r-\boldsymbol{x}_s\vert \vert ^2\right) +\\
\beta_2^{-1} \delta(\boldsymbol{x}_r,\boldsymbol{x}_s)\text{,}
\end{multline}
with parameters $\beta_1$ and $\beta_2$ (with $\delta(\boldsymbol{x}_r,\boldsymbol{x}_s)$ we indicate the Kronecker delta). 
Instead here, we adopt a richer ARD structure for the SE kernel, characterized by a different length-scale for each latent state component:
\begin{multline}\label{eq: my y kernel}
k_y(\boldsymbol{x}_r, \boldsymbol{x}_s) = \text{exp}\left(-\vert \vert \boldsymbol{x}_r-\boldsymbol{x}_s\vert \vert _{\mathbf{\Lambda}_y^{-1}}\right) +\\
\sigma_y^2 \delta(\boldsymbol{x}_r,\boldsymbol{x}_s)\text{.}
\end{multline}
$\mathbf{\Lambda}_y^{-1} = \text{diag}(\lambda_{y,1}^{-2},\dots,\lambda_{y,D}^{-2})$ is a positive definite diagonal matrix, which weights the norm used in the SE function, and $\sigma_y^2$ is the variance of the isotropic noise in \eqref{eq: latent mapping}. The trainable hyper-parameters of the latent map model are then $\boldsymbol{\theta}_y = \left[w_{y,1},\dots, w_{y,D}, \lambda_{y,1},\dots,\lambda_{y,D}, \sigma_y\right]^T$. 

\subsubsection{Dynamics mapping}\label{subsubsec:dynamic map}
Similarly to Sec.~\ref{subsubsec:latent map}, we can model a priori each component of the latent state difference  $\boldsymbol{x}_{t+1}-\boldsymbol{x}_t = [x_{t+1}^{(1)}-x_t^{(1)}, \dots, x_{t+1}^{(d)}-x_t^{(d)} ]^T$ as a zero-mean GP that takes as input the pair $(\boldsymbol{x}_t,\boldsymbol{u}_t)$, for $t=1,\dots,N-1$.

Let $\mathbf{X} = [ \boldsymbol{x}_1,\dots, \boldsymbol{x}_N]^T \in \mathbb{R}^{N\times d}$ be the matrix collecting the set of $N$ latent states, we can denote by $\mathbf{X}_{r:s,i}$ the vector of the $i$-th components from time step $r$ to time step $s$, with $r,s=1,\dots,N$. We indicate the vector of differences between consecutive latent states along their $i$-th component with $\mathbf{\Delta}_{:,i} = (\mathbf{X}_{2:N,i} - \mathbf{X}_{1:N-1,i})\in \mathbb{R}^{N-1}$. $\mathbf{\Delta} = [\mathbf{\Delta}_{:,1},\dots,\mathbf{\Delta}_{:,d} ]\in \mathbb{R}^{(N-1)\times d}$ is the matrix that collects differences along all the components.

Finally, we compactly represent the GP input of the dynamic model as $\tilde{\boldsymbol{x}}_t = [\boldsymbol{x}_t^T, \boldsymbol{u}_t^T]^T \in \mathbb{R}^{d+E}$, and refer to the the matrix collecting $\tilde{\boldsymbol{x}}_t$ for $t=1,\dots,N-1$ with $\tilde{\mathbf{X}} = \left[ \tilde{\boldsymbol{x}}_1,\dots, \tilde{\boldsymbol{x}}_{N-1}\right]^T \in\mathbb{R}^{(N-1) \times (d+E)}$.
With similar assumptions to the ones made for the latent map, and denoting the common kernel function for all the GPs with $k_x(\cdot,\cdot)$, and the different scaling factors with $w_{x,i}$, for $i=1,\dots,d$, the joint likelihood is given by
\begin{multline}\label{eq: dynamic simplified density}
    p(\mathbf{\Delta}\vert \tilde{\mathbf{X}}) = \frac{\vert \mathbf{W}_x\vert ^{N-1}}{\sqrt{(2\pi)^{(N-1)d}\vert \mathbf{K}_x(\tilde{\mathbf{X}})\vert ^d}} \cdot\\
     \text{exp}\left( -\frac{1}{2} \text{tr}\left(\left(\mathbf{K}_x(\tilde{\mathbf{X}})\right)^{-1} \mathbf{\Delta} \mathbf{W}_x^2 \mathbf{\Delta}^T\right)\right)\text{,}
\end{multline}
where $\mathbf{W}_x=\text{diag}(w_{x,1},\dots,w_{x,d})$ and $\mathbf{K}_x(\tilde{\mathbf{X}})$ is the covariance matrix defined by  $k_x(\cdot,\cdot)$.
In standard GPDM \cite{wang2005gaussian}, dynamic mapping GPs have been proposed with constant scaling factors $w_{x,i}=1$ for $i=1,\dots,d$, and equipped with a naive kernel resulting from the sum of an isotrophic SE and an homogeneous linear function, with only four trainable parameters:
\begin{multline}\label{eq: old x kernel}
k_x'(\tilde{\boldsymbol{x}}_r, \tilde{\boldsymbol{x}}_s) = \alpha_1\text{exp}\left(-\frac{\alpha_2}{2}\vert \vert \tilde{\boldsymbol{x}}_r-\tilde{\boldsymbol{x}}_s\vert \vert ^2\right)+ \\
\alpha_3 \tilde{\boldsymbol{x}}_r^T\tilde{\boldsymbol{x}}_s + \alpha_4^{-1} \delta(\tilde{\boldsymbol{x}}_r,\tilde{\boldsymbol{x}}_s)\text{.}
\end{multline}

Analogously to the latent mapping case, we decided to adopt the following kernel function,
\begin{multline}\label{eq: my x kernel}
k_x(\tilde{\boldsymbol{x}}_r, \tilde{\boldsymbol{x}}_s) = \text{exp}\left(-\vert \vert \tilde{\boldsymbol{x}}_r-\tilde{\boldsymbol{x}}_s\vert \vert _{\mathbf{\Lambda}_{x}^{-1}}\right) +\\
[\tilde{\boldsymbol{x}}_r^T, 1]\mathbf{\Phi} [\tilde{\boldsymbol{x}}_s^T 1]^T + \sigma_x^2\delta(\tilde{\boldsymbol{x}}_r,\tilde{\boldsymbol{x}}_s)    \text{.}
\end{multline}
$\mathbf{\Lambda}_{x}^{-1} = \text{diag}(\lambda_{x,1}^{-2},\dots,\lambda_{x,d+E}^{-2})$ is a positive definite diagonal matrix, which weights the norm used in the SE component of the kernel.
Also $\mathbf{\Phi} = \text{diag}(\phi_{1}^2,\dots,\phi_{d+E+1}^2) $ is a positive definite diagonal matrix that describes the linear component. $\sigma_x^2$ is the variance of the isotropic noise in \eqref{eq: markov dynamics}.
In comparison to \eqref{eq: old x kernel}, the adopted kernel weights differently the various components of the input in both SE and linear part, where the GP input is also extended as $\left[\tilde{\boldsymbol{x}}_s^T, 1\right]^T$.
The trainable hyper-parameters of the dynamic map model are then $\boldsymbol{\theta}_x = \left[w_{x,1},\dots, w_{x,d}, \lambda_{x,1},\dots,\lambda_{x,d}, \phi_{1},\dots,\phi_{d+E+1}, \sigma_x\right]^T$. 

In the following, we will refer with \textit{naive} CGPDM to the model that straightforwardly extends the standard GPDM structure from \cite{wang2005gaussian}, using its same kernels, \eqref{eq: old y kernel},\eqref{eq: old x kernel}, and constant scaling factors;
while we denote with \textit{advanced} CGPDM the proposed model characterized by kernels \eqref{eq: my y kernel},\eqref{eq: my x kernel} and trainable scaling factors in the dynamical map.
Although ARD kernels are commonly adopted in GP regression \cite{williams2006gaussian}, they were not tested before in GPDMs.
Trainable scaling factors constitute a novelty for this kind of model too.

\subsubsection{Working with multiple sequences}\label{sec:multiple sequences}
It is possible to easily extend  the CGPDM formulation to $P$ multiple sequences of observations, $\mathbf{Y}^{(1)}, \dots, \mathbf{Y}^{(P)}$, and control inputs, $\mathbf{U}^{(1)}, \dots, \mathbf{U}^{(P)}$.
Let the length of each sequence $p$, for $p=1,\dots,P$, be equal to $N_p$, with $\sum_{p=1}^PN_p = N$.
Define the latent states associated with each sequence as $\mathbf{X}^{(1)}, \dots, \mathbf{X}^{(P)}$.
Following the notation of Sec.~\ref{subsubsec:dynamic map}, define $\tilde{\mathbf{X}}^{(1)}, \dots, \tilde{\mathbf{X}}^{(P)}$, as the sequence of the aggregated matrices of latent states and control inputs, and  $\mathbf{\Delta}^{(1)}, \dots, \mathbf{\Delta}^{(P)}$ as the difference matrices.
Hence, model joint likelihoods can be calculated by using the following concatenated matrices inside \eqref{eq: latent simplified density} and \eqref{eq: dynamic simplified density}: $\mathbf{Y} = [\mathbf{Y}^{(1)T}\vert  \dots \vert  \mathbf{Y}^{(P)T}]^T$, $\mathbf{X} = [\mathbf{X}^{(1)T}\vert  \dots \vert  \mathbf{X}^{(P)T}]^T$, $\mathbf{\Delta} = [\mathbf{\Delta}^{(1)T}\vert  \dots\vert  \mathbf{\Delta}^{(P)T}]^T$ and $\tilde{\mathbf{X}} = [\tilde{\mathbf{X}}^{(1)T}\vert  \dots\vert  \tilde{\mathbf{X}}^{(P)T}]^T$.
Note that, when dealing with multiple sequences, the number of data points in the dynamic mapping becomes $N-P$, and expression \eqref{eq: dynamic simplified density} must change accordingly.

\subsection{CGPDM Training and Prediction}\label{subsec:CGPDM learning and pred}
\begin{figure*}[tpb]
	\centering
	\includegraphics[width=\textwidth , keepaspectratio]{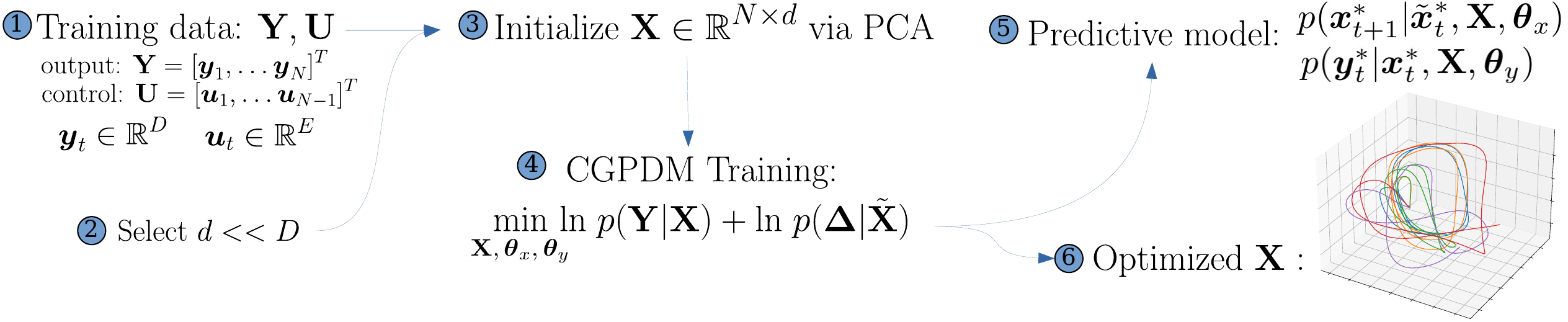}
	\caption{
    Flowchart summarizing the CGPDM training process. Given a set of training data \textbf{Y} and \textbf{U} (1) and a desired latent dimension $d$ (2), the associated latent states \textbf{X} are initialized via PCA (3) and then optimized together with other CGPDM hyper-parameters (4). After the training, we obtain the probabilistic predictive model (5), and the set of optimized latent trajectories capturing the high-dimensional dynamics (6)
    }
	\label{fig:flowchart}
\end{figure*}
Training a CGPDM entails using numerical optimization techniques to estimate the unknowns in the model, i.e., latent states $\mathbf{X}$ and the hyper-parameters $\boldsymbol{\theta}_x,\boldsymbol{\theta}_y$.
Latent coordinates $\mathbf{X}$ are initialized by means of PCA \cite{bishop2006pattern}, selecting the first $d$ principal components of $\mathbf{Y}$. A natural approach for training CGPDMs is to maximize the joint log-likelihood $\text{ln}\;p(\mathbf{Y}\vert \mathbf{X}) +\text{ln}\;p(\mathbf{\Delta}\vert \tilde{\mathbf{X}})$ w.r.t. $\{\mathbf{X}, \boldsymbol{\theta}_x,\boldsymbol{\theta}_y\}$.
To do so, in this work, we adopted the L-BFGS algorithm \cite{byrd1995limited}.

The overall loss to be optimized can be written as $\mathcal{L} = \mathcal{L}_y+ \mathcal{L}_x$, with $\mathcal{L}_y$ and $\mathcal{L}_x$ defined as
\begin{multline}
    \mathcal{L}_y = \frac{D}{2}\text{ln}\vert \mathbf{K}_y(\mathbf{X})\vert  + \frac{1}{2}\text{tr}(\mathbf{K}_y(\mathbf{X})^{-1}\mathbf{Y} \mathbf{W}_y^2 \mathbf{Y}^T)-\\
    N \text{ln} \vert \mathbf{W}_y\vert \text{,}
\label{eq:loss_y}
\end{multline}
\begin{multline}
    \mathcal{L}_x = \frac{d}{2}\text{ln}\vert \mathbf{K}_x(\tilde{\mathbf{X}})\vert  + \frac{1}{2}\text{tr}(\mathbf{K}_x(\tilde{\mathbf{X}})^{-1} \mathbf{\Delta} \mathbf{W}_x^2 \mathbf{\Delta}^T)-\\
    (N-1) \text{ln} \vert \mathbf{W}_x\vert \text{.}
\label{eq:loss_x}
\end{multline}

In case the CGPDM is trained on multiple sequences of inputs and observations, make sure to employ the aggregated matrices defined in Sec.~\ref{sec:multiple sequences} when computing loss functions \ref{eq:loss_y}-\ref{eq:loss_x}.
It is also necessary to use the factor $N-P$ instead of $N-1$ inside the $\mathcal{L}_x$ expression.
The overall training procedure is represented schematically in Fig. \ref{fig:flowchart}.

A trained CGPDM can be used to fulfill two different purposes: (i) map a given new latent state $\boldsymbol{x}_t^*$ to the corresponding $\boldsymbol{y}_t^*$ in observation space, (ii) predict the evolution of the latent state at the next time step $\boldsymbol{x}_{t+1}^*$, given $\boldsymbol{x}_{t}^*$ and a certain control $\boldsymbol{u}_{t}^*$. The two processes, together, can  predict the observations produced by a given series of control actions.

\subsubsection{Latent prediction}\label{subsubsec:latent pred}
Given $\boldsymbol{x}_t^*$, its corresponding $\boldsymbol{y}_t^*$ is distributed as $p(\boldsymbol{y}_t^*\vert \boldsymbol{x}_t^*, \mathbf{X}, \boldsymbol{\theta}_y) = \mathcal{N}(\boldsymbol{\mu}_y(\boldsymbol{x}_t^*),v_y(\boldsymbol{x}_t^*)\mathbf{W}_y^{-2})$, with
\begin{equation}
    \boldsymbol{\mu}_y(\boldsymbol{x}_t^*) = \mathbf{Y}^T \mathbf{K}_y(\mathbf{X})^{-1} \boldsymbol{k}_y(\boldsymbol{x}_t^*,\mathbf{X})
\end{equation}
\begin{multline}
    v_y(\boldsymbol{x}_t^*) = k_y(\boldsymbol{x}_t^*,\boldsymbol{x}_t^*) -\\
    \boldsymbol{k}_y(\boldsymbol{x}_t^*,\mathbf{X})^T \mathbf{K}_y(\mathbf{X})^{-1} \boldsymbol{k}_y(\boldsymbol{x}_t^*,\mathbf{X}) \text{,}
\end{multline}
where $\boldsymbol{k}_y(\boldsymbol{x}_t^*,\mathbf{X}) = \left[k_y(\boldsymbol{x}_t^*,\boldsymbol{x}_1),\dots,  k_y(\boldsymbol{x}_t^*,\boldsymbol{x}_N)\right]^T$.

\subsubsection{Dynamics prediction}\label{subsubsec:dyn pred}
Given $\boldsymbol{x}_t^*$ and $\boldsymbol{u}_t^*$, let's define $\tilde{\boldsymbol{x}}_t^*=[\boldsymbol{x}_t^{*T},\boldsymbol{u}_t^{*T}]^T$. The probability density of  the latent state at the next time step $\boldsymbol{x}_{t+1}^*$ is $p(\boldsymbol{x}_{t+1}^*\vert \tilde{\boldsymbol{x}}_t^*, \mathbf{X}, \boldsymbol{\theta}_x) = \mathcal{N}(\boldsymbol{\mu}_x(\boldsymbol{x}_t^*),v_x(\boldsymbol{x}_t^*)\mathbf{W}_x^{-2})$, with
\begin{equation}
    \boldsymbol{\mu}_x(\boldsymbol{x}_t^*) = \boldsymbol{x}_t^* + \mathbf{\Delta}^T \mathbf{K}_x(\tilde{\mathbf{X}})^{-1} \boldsymbol{k}_x(\tilde{\boldsymbol{x}}_t^*,\tilde{\mathbf{X}})\text{,}
\end{equation}
\begin{multline}
    v_x(\boldsymbol{x}_t^*) = k_x(\tilde{\boldsymbol{x}}_t^*,\tilde{\boldsymbol{x}}_t^*) -\\
    \boldsymbol{k}_x(\tilde{\boldsymbol{x}}_t^*,\tilde{\mathbf{X}})^T \mathbf{K}_x(\tilde{\mathbf{X}})^{-1} \boldsymbol{k}_x(\tilde{\boldsymbol{x}}_t^*,\tilde{\mathbf{X}})\text{,}
\end{multline}
with $\boldsymbol{k}_x(\tilde{\boldsymbol{x}}_t^*,\tilde{\mathbf{X}})=\left[k_x(\tilde{\boldsymbol{x}}_t^*,\tilde{\boldsymbol{x}}_1)\dots  k_x(\tilde{\boldsymbol{x}}_t^*,\tilde{\boldsymbol{x}}_{N-1})\right]^T$.

\subsubsection{Trajectory prediction}\label{subsubsec:trajectory prediction}
Starting from an initial latent state $\boldsymbol{x}_1^*$, one can predict the system evolution over a desired horizon of length $N_d$, when subject to a given sequence of control actions $\boldsymbol{u}_1^*,\dots,\boldsymbol{u}_{N_d-1}^*$. At each time step $t=1,\dots,N_d-1$, $\boldsymbol{x}_{t+1}^*$ can be sampled from the normal distribution $p(\boldsymbol{x}_{t+1}^*\vert \tilde{\boldsymbol{x}}_t^*, \mathbf{X}, \boldsymbol{\theta}_x)$ defined in Sec.~\ref{subsubsec:dyn pred}. Hence, the generated trajectory in the latent space $\boldsymbol{x}_1^*,\dots,\boldsymbol{x}_{N_d}^*$ can be mapped into the associated sequences of observations $\boldsymbol{y}_1^*,\dots,\boldsymbol{y}_{N_d}^*$ by considering the previously defined probability distribution $p(\boldsymbol{y}_t^*\vert \boldsymbol{x}_t^*, \mathbf{X}, \boldsymbol{\theta}_y)$.

\section{Results}\label{sec:results}
We employed the proposed CGPDM to model the high-dimensional dynamics that characterizes the motion of a piece of cloth held by a robotic system.
This section reports the results obtained in two sets of experiments: a simulated session (Subsec.~\ref{subsec:sim_experiment}) and one conducted on a real setup (Subsec.~\ref{subsec:real_experiment}).
We exploited simulation to assess the performance of CGPDM over a wide set of scenarios (different amount of training data, motion ranges, and model structure), while the real-world experiment served as validation over non-synthetic data.
The objective of the experiments was to learn the high-dimensional cloth dynamics using CGPDM, in order to make predictions about cloth movements in response to sequences of actions that were not seen during training.
In particular, we aimed to evaluate how  model prediction accuracy is affected by:
\begin{itemize}
    \item the number of data used for training,
    \item the oscillation range of the cloth movements,
    \item the use of \textit{advanced} or \textit{naive} CGPDM structures (as defined in Sec.~\ref{subsec:CGPDM}).
\end{itemize}
Such high-dimensional task would be unfeasible to model by standard GP regression without DR. CGPDMs were implemented in Python\footnote[1]{Code publicly available at \url{https://github.com/fabio-amadio/cgpdm_lib}}, employing PyTorch \cite{paszke2019pytorch}.

\subsection{Simulated Cloth Experiment}\label{subsec:sim_experiment}
\begin{figure}[t]
\centering
\includegraphics[width=0.95\linewidth]{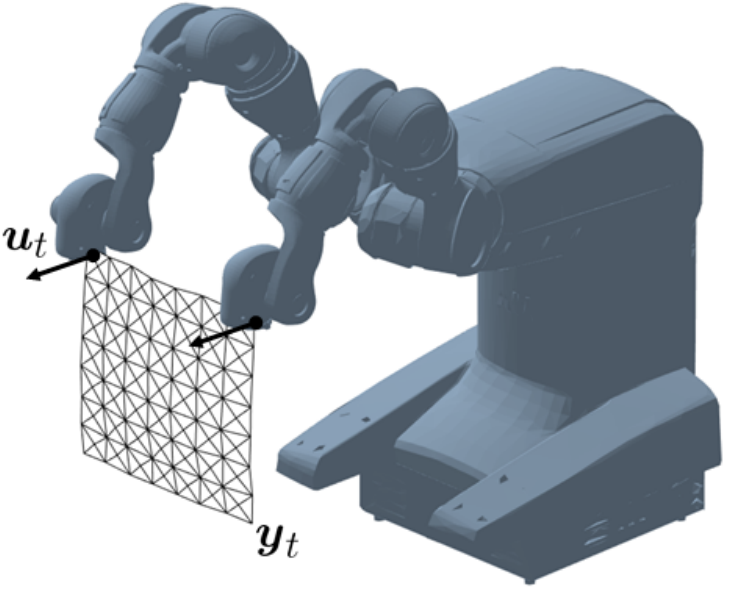}
\caption{Simulated setup for cloth manipulation with bimanual robot. The cloth is positioned in its starting configuration}
\label{fig:sim_setup}
\end{figure}

In the simulated scenario, we considered a bimanual robot moving a squared piece of cloth by holding its two upper corners, as shown in Fig. \ref{fig:sim_setup}.
The cloth was modeled as an 8$\times$8 mesh of material points.
We made the assumption that the two upper corner points are attached to the robot's end-effectors, while the other points move freely following the dynamical model proposed in \cite{francoModel}.

In this context, the observation vector is given by the Cartesian coordinates of all the points in the mesh (measured in meters); hence $\boldsymbol{y}_t\in\mathbb{R}^D$ with $D=192$.
We assumed to control exactly the two robot arms in the operational space, keeping the same orientation and relative distance between the two end-effectors and producing oscillation in the Y-Z plane.
Thus, we considered as control actions the differences between consecutive commanded end-effector positions in the Y and Z directions, resulting in a $\boldsymbol{u}_t\in\mathbb{R}^E$ with $E=2$.

\subsubsection{Data collection}
Training and test data were obtained by recording mesh trajectories associated with several types of cloth oscillation, obtained by applying different sequences of control actions. All the considered trajectories start from the same cloth configuration and last 5 seconds. Observations were recorded at 20 Hz, hence $N=100$ total number of steps for each sequence.

Robot end-effectors move in a coordinate fashion drawing oscillations on the Y-Z plane.
Let $\boldsymbol{u}_t = \left[\Delta ee^Y_t, \Delta ee^Z_t\right]^T$, where $\Delta ee^Y_t,$ and $\Delta ee^Z_t,$ indicate the difference between consecutive end-effector commanded positions along the Y and Z axes. Specifically, their values were given by the two following periodic expressions:
\begin{multline}
    \Delta ee^{Y}_t=A\cdot \textrm{cos}(2\pi f_{Y} t) \left[-\textrm{cos}(\gamma),\textrm{sin}(\gamma)\right]\text{,}\\
     \Delta ee^{Z}_t=A\cdot \textrm{cos}(2\pi f_{Z} t) \left[-\textrm{cos}(\gamma),\textrm{sin}(\gamma)\right]\text{.}
\label{eq: oscillation control}
\end{multline}

Such controls make the end-effectors oscillate on the Y-Z plane of the operational space.
The maximum displacement is regulated by $A$, that we set to 0.01 meters.
Parameter $\gamma$ can be interpreted as the inclination of $\boldsymbol{u}_1$ w.r.t. the horizontal, and it loosely defines a direction of the oscillation. $f_Y$ and $f_Z$ define the frequencies of the oscillations along Y and Z axes.
If they are similar, the end-effectors move mostly along the direction defined by $\gamma$, if not, they swipe in a broader space. 

\begin{figure}[t]
\centering
\includegraphics[width=0.95\linewidth]{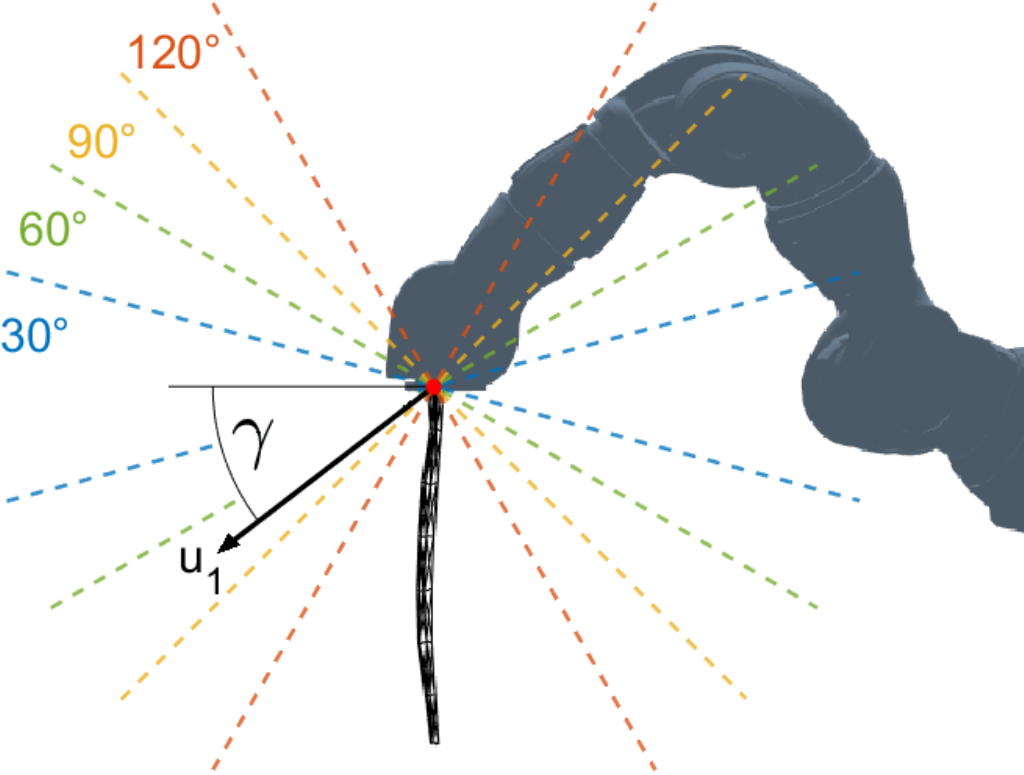}
\caption{Oscillation ranges $R\in\{30\text{°},60\text{°},90\text{°},120\text{°}\}$ defining the sampling intervals for $\gamma$ during data collection}
\label{fig:ranges}
\end{figure}

In order to obtain a heterogeneous set of trajectories for the composition of training and test sets, we collected several movements obtained by choosing in a random fashion the control parameters $\gamma$, $f_Y$ and $f_Z$. Angles $\gamma$ were uniformly sampled inside a variable range $[-\frac{R}{2},\frac{R}{2}]$ (deg); in the following, we indicate this range with the amplitude of its angular area, $R$ (deg). Instead, frequencies $f_Y$ and $f_Z$ were uniformly sampled inside the fixed interval [0.3, 0.6] (Hz). 
We considered four movement ranges of increasing width, namely $R\in\{30\text{°},60\text{°},90\text{°},120\text{°}\}$ (Fig. \ref{fig:ranges}), and collected a specific data-set $\mathcal{D}_R$ associated with each range. Every set contains 50 cloth trajectories obtained by applying control actions of the form \eqref{eq: oscillation control} with 50 different random choices for parameters $\gamma$, $f_Y$ and $f_Z$. From each $\mathcal{D}_R$, 10 trajectories were extracted and used as test sets $\mathcal{D}_R^{test}$ for the corresponding movement range, while several training sets $\mathcal{D}_R^{train}$ were built by randomly picking from the remaining sequences.

\begin{figure*}[tpb]
	\centering
	\includegraphics[width=\textwidth , keepaspectratio]{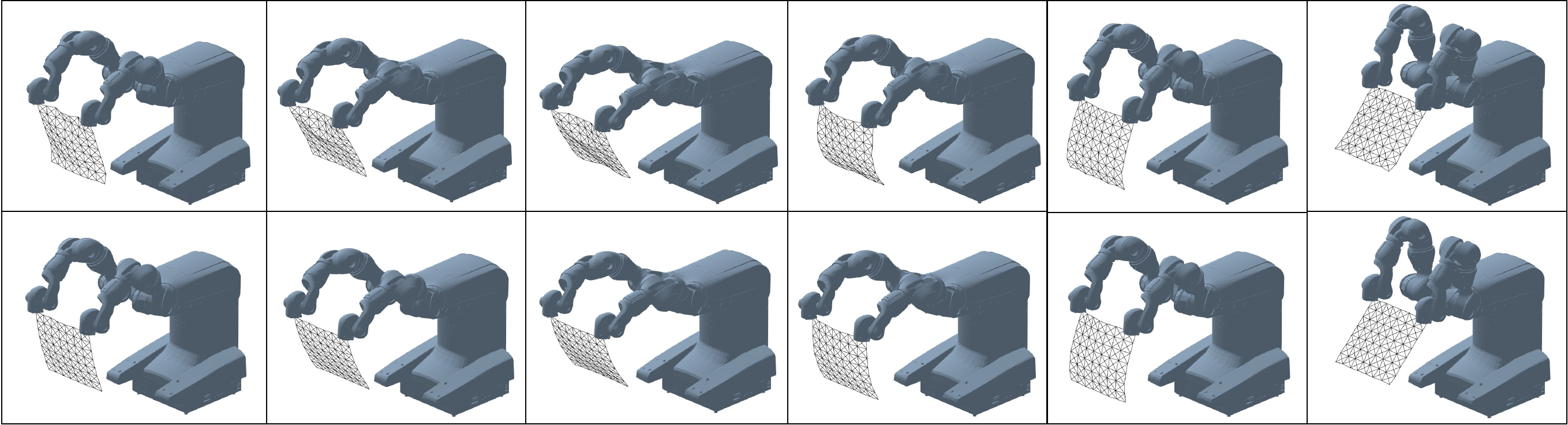}
	\caption{True (top) and predicted (bottom) simulated cloth oscillation frames for one of the considered test trajectories}
	\label{fig:time_frames}
\end{figure*}

\subsubsection{Model training}\label{subsubsec:training_models}
In all the models, we adopted a latent space of dimension $d=3$, resulting in a dimensionality reduction factor of $D/d=64$.
This $d$ value was chosen empirically after preliminary tests and allows to easily visualize the latent variables behaviour in a three-dimensional space, see for instance Fig. \ref{fig:latent mapping}.
Other choices are possible, but such sensitivity analysis is left out of the scope of this experimental analysis.

The objective of the experiment was to evaluate CGPDM prediction accuracy at different movement ranges, and for different amounts of training data.
Moreover, we wanted to observe if the use of the proposed \textit{advanced} CGPDM structure yields a substantial difference in terms of accuracy when compared to the \textit{naive} model.
Consequently, for each considered movement range $R$, we trained two different sets of CGPDMs, adopting in one the \textit{naive} structure and in the other the \textit{advanced} one.
Each model in the two sets was trained employing an increasing number of sequences randomly picked from $\mathcal{D}_R^{train}$.
Specifically, we used 5 different random combinations of 5, 10, 15 and 20 sequences for each oscillation range (varying each time the random seed).
In this way, we were able to reduce the dependencies on the specific training trajectories considered, and to average prediction accuracy over different possible sets of training data.

\subsubsection{Model prediction}\label{subsubsec:sim_exp_results}
We used each learned CGPDM to predict the cloth movements when subject to the control actions observed for each test sequence inside $\mathcal{D}_R^{test}$, with $R\in\{30\text{°},60\text{°},90\text{°},120\text{°}\}$.
Let $\boldsymbol{y}_t^{(R,k)}$ and $\boldsymbol{u}_t^{(R,k)}$ denote, respectively, the observation and control action at time step $t$ of the $k$-th test trajectory in $\mathcal{D}_R^{test}$ (with $k=1,\dots,10$).

For each considered range $R$, one can follow the procedure of Sec.~\ref{subsubsec:trajectory prediction} and employ the trained CGPDMs to predict the trajectories resulting from the application of $\{\boldsymbol{u}_t^{(R,k)}\}_{t=1}^{N-1}$, for $k=1,\dots,10$.
Let $\boldsymbol{x}_t^{*(R,k)}$ be the predicted latent state at time $t$, and $\boldsymbol{y}_t^{*(R,k)}$ the corresponding predicted observation.
As an example, in Fig. \ref{fig:time_frames} we show a sequence of true and predicted cloth configurations for one of the considered test trajectory.
Please, refer to the video\footnote[2]{\label{note1}Videos of the experiments (simulated and real) are available at \url{https://youtu.be/JnqkelnP5-E}} for a clearer visualization of the obtained results.

\begin{figure*}[ht]
	\centering
	\includegraphics[width=\textwidth , keepaspectratio]{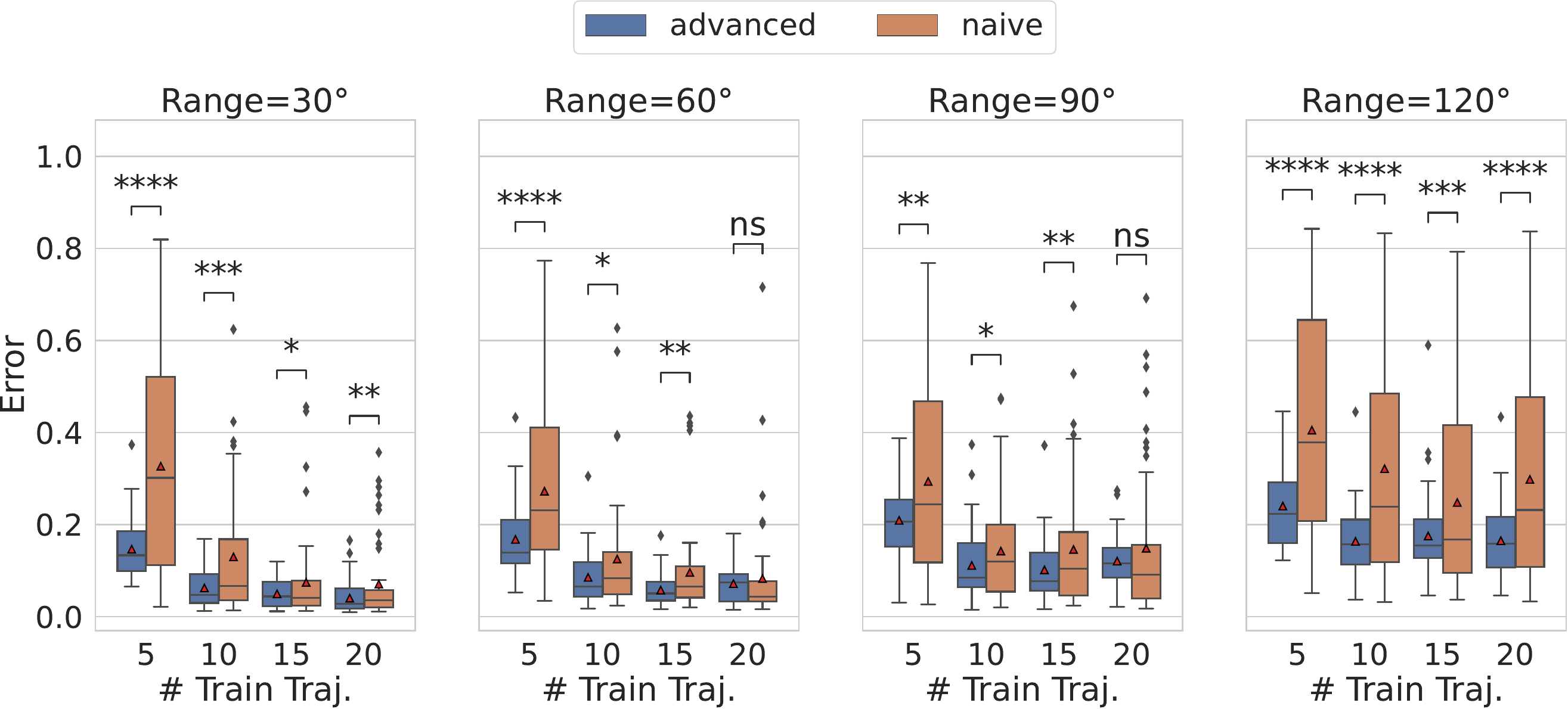}
	\caption{Boxplot representing the test prediction errors obtained by the \textit{advanced} and the \textit{naive} CGPDM structures at different oscillation ranges in the simulation experiment. Each configuration was tested with 5 different randomly composed $\mathcal{D}_R^{train}$. Mean values are indicated with red triangles and statistical  significance of T-test comparing \textit{advanced} and \textit{naive} results are represented with the following notation: $\texttt{ns}: 5.0\mathrm{e}{-2}< p \leq 1.0$, $*: 1.0\mathrm{e}{-2} < p \leq 5.0\mathrm{e}{-2}$, $**: 1.0\mathrm{e}{-3} < p \leq 1.0\mathrm{e}{-2}$, $***: 1.0\mathrm{e}{-4} < p \leq 1.0\mathrm{e}{-3}$, $****: p \leq 1.0\mathrm{e}{-4}$)
    }
	\label{fig:boxplot}
\end{figure*}

For every predicted trajectory, we measured the average distance between the real and the predicted mesh points.
Fig. \ref{fig:boxplot} represents the observed errors by means of boxplots, indicating also the statistical relevance of the \textit{naive}-\textit{advanced} difference in each experiment configuration (T-test performed by using the open-source library Statannotations\footnote[3]{Statannotations library available at \url{https://github.com/trevismd/statannotations}}).
Moreover, Table \ref{tab:errors} reports the average distances between true and predicted mesh points obtained in the test sets by the different CGPDM configurations in all the movement ranges.
Results are expressed in terms of mean and 95\% confidence intervals obtained by averaging over the different training sets adopted (all the experiments were repeated 5 times, using a randomly composed $\mathcal{D}_R^{train}$).

\begin{table*}[ht]
\footnotesize
\caption{Mean prediction errors (with 95\% C.I.) between true and predicted cloth trajectories obtained by the \textit{advanced} and the \textit{naive} CGPDM structures at different oscillation ranges in both the simulated and real-world experiment (the number of data used for training is indicated inside the squared brackets)}
\begin{center}
\begin{tabular}{r*{8}{c}}
    \hline
    \multirow{2}{*}{Experiment} & \multicolumn{2}{c}{$R=30$°} & \multicolumn{2}{c}{$R=60$°} & \multicolumn{2}{c}{$R=90$°} & \multicolumn{2}{c}{$R=120$°}\\
    \cline{2-9}
    & \textit{advanced} & \textit{naive} & \textit{advanced} & \textit{naive} & \textit{advanced} & \textit{naive} & \textit{advanced} & \textit{naive}\\
    \hline
    Sim [5]  & $0.15 \pm 0.02$ & $0.33 \pm 0.06$ & $0.17 \pm 0.02$ & $0.27 \pm 0.05$ & $0.21 \pm 0.02$ & $0.29 \pm 0.06$ & $0.24 \pm 0.03$ & $0.41 \pm 0.07$\\
    \hline
    Sim [10] & $0.06 \pm 0.01$ & $0.13 \pm 0.04$ & $0.09 \pm 0.02$ & $0.13 \pm 0.03$ & $0.11 \pm 0.02$ & $0.14 \pm 0.03$ & $0.16 \pm 0.02$ & $0.32 \pm 0.06$\\
    \hline
    Sim [15] & $0.05 \pm 0.01$ & $0.07 \pm 0.03$ & $0.06 \pm 0.01$ & $0.10 \pm 0.03$ & $0.10 \pm 0.02$ & $0.15 \pm 0.04$ & $0.18 \pm 0.02$ & $0.25 \pm 0.06$\\
    \hline
    Sim [20] & $0.04 \pm 0.01$ & $0.07 \pm 0.02$ & $0.07 \pm 0.01$ & $0.08 \pm 0.03$ & $0.12 \pm 0.02$ & $0.15 \pm 0.04$ & $0.17 \pm 0.02$ & $0.30 \pm 0.06$\\
    \hline
    \hline
    Real [9] & $0.11 \pm 0.02$ & $0.18 \pm 0.06$ & $0.09 \pm 0.01$ & $0.22 \pm 0.07$ & $0.14 \pm 0.04$ & $0.31 \pm 0.03$ & $ND$ & $ND$\\

\end{tabular}
\end{center}
\label{tab:errors}
\end{table*}

\subsection{Real Cloth Experiment}\label{subsec:real_experiment}
\begin{figure}[t]
\centering
\includegraphics[width=0.9\linewidth]{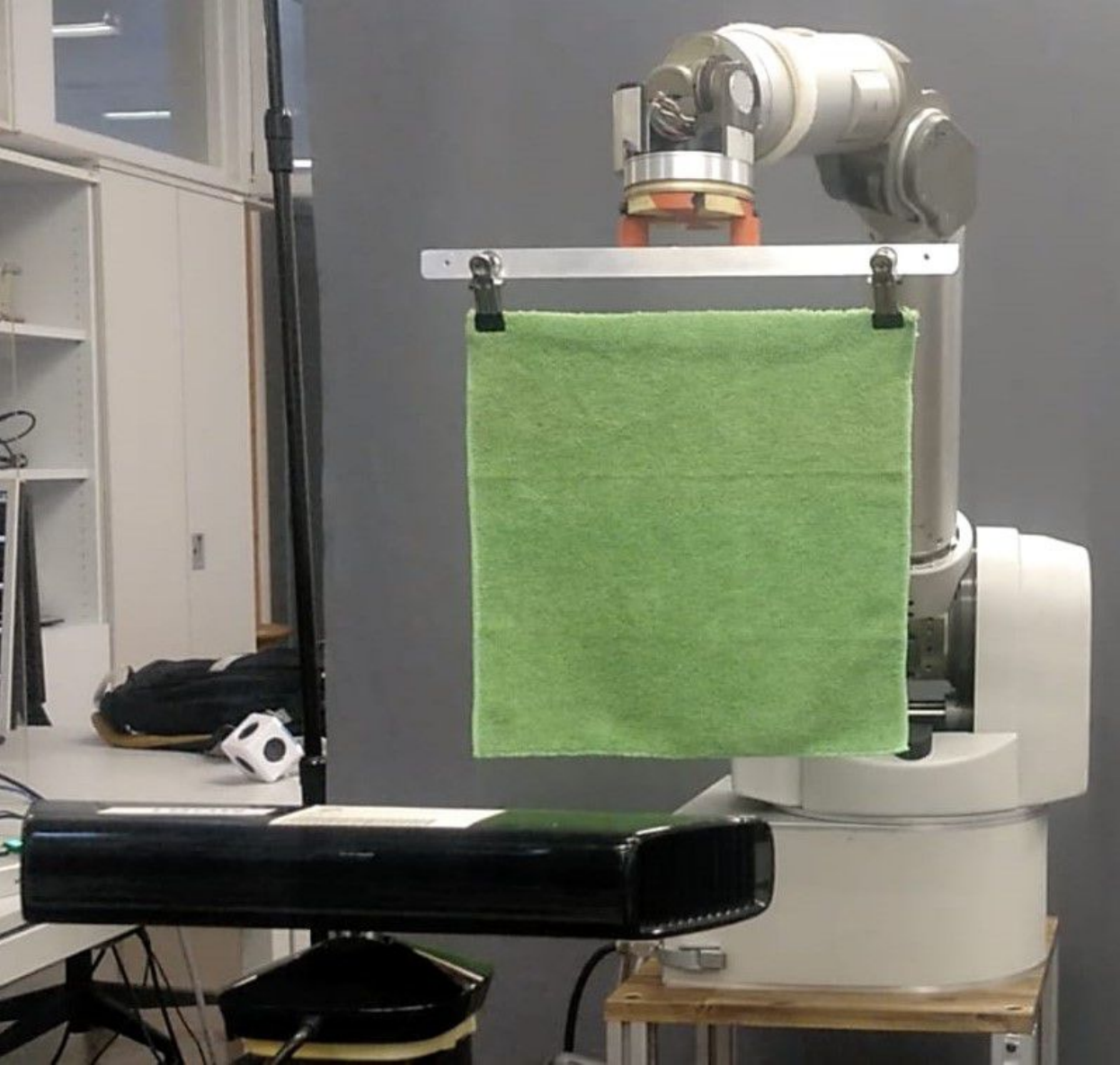}
\caption{Real experimental setup with the Barrett WAM Arm holding a piece of cloth whose motion can be tracked by a RGBD camera}
\label{fig:real_setup}
\end{figure}

In this second set of experiments, we tested CGPDM on data collected in a real cloth manipulation scenario.
For this purpose, we used a Barrett WAM Arm\footnote[4]{Barrett WAM Arm: \url{https://advanced.barrett.com/wam-arm-1}}, whose end-effector consists of a coat rack that can firmly grip a piece of cloth from its corners.
The overall setup is depicted in Fig. \ref{fig:real_setup}.
We controlled the robot's end-effector in position, recording the resulting movement of the cloth through a motion capture system based on information extracted from an RGBD camera.
We combined object detection, image and point cloud processing for segmenting cloth-like objects\footnote[5]{Code publicly available at \url{https://github.com/MiguelARD/cloth_point_cloud_segmentation}}, following \cite{bochkovskiy2020yolov4}, \cite{GrabCut} and \cite{colorSegmentation}.

\subsubsection{Data Collection}
As in the simulated scenario, we captured the cloth as an 8$\times$8 mesh of points, whose spatial coordinates constitute the observation vector $\boldsymbol{y}_t\in\mathbb{R}^D$ with $D=192$.

Control actions were defined following again expressions \eqref{eq: oscillation control} and commanded to the robot at 100 Hz.
Parameters $f_Y$ and $f_Z$ were uniformly sampled within [0.2, 0.5] (Hz) and A was set to 0.004 meters.
In this experiment, we considered only the $R=30$°, $R=60$°, and $R=90$° oscillation ranges ($R=120$° was excluded because of robot workspace limitations).

The motion capture system could work only at rates lower than 100 Hz, with no guaranteed sampling interval length.
Thus, it was necessary to post-process the data to make them ready for modeling.
Firstly, motion capture data were smoothed by a moving average filter.
Then we interpolated the positions of both the end-effector and the cloth mesh, to obtain two synchronized sequences of observations and control actions, sampled at 20 Hz.
For each of the three ranges, we collected 10 trajectories each 3 seconds long.

\subsubsection{Model training \& prediction}
For every considered oscillation range, we trained two sets of CGPDMs, one using the \textit{naive} and one the \textit{advanced} model structure.
Each set of trajectories is composed of 10 sequences, hence we followed a cross-validation method for training and testing the models.
At every range, we trained the models using all the sequences but one, left out for testing, repeating the procedure ten times varying the test sequence each time.

\begin{figure}[t]
\centering
\includegraphics[width=\linewidth]{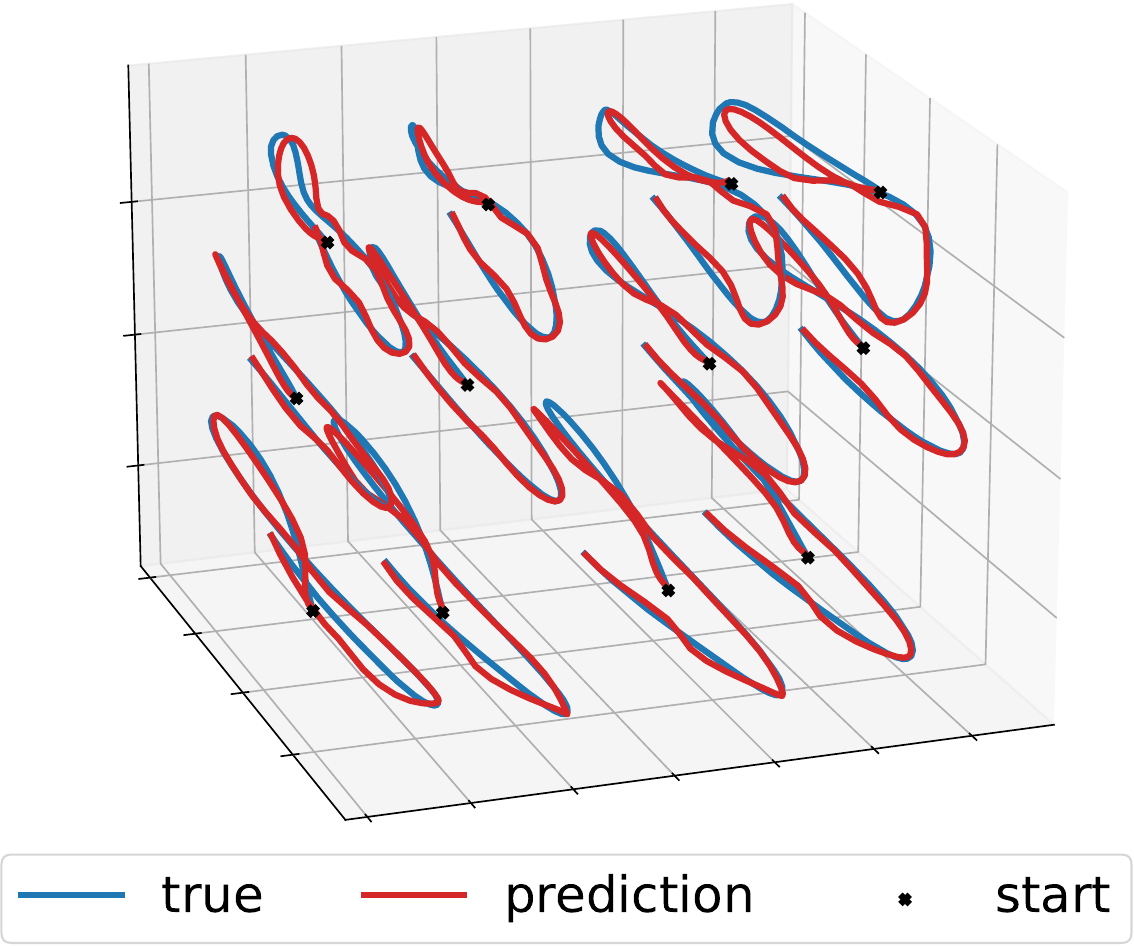}
\caption{True and predicted mesh points for one of the registered real cloth movement}
\label{fig:real rollout}
\end{figure}

\begin{figure}[t]
\centering
\includegraphics[width=\linewidth]{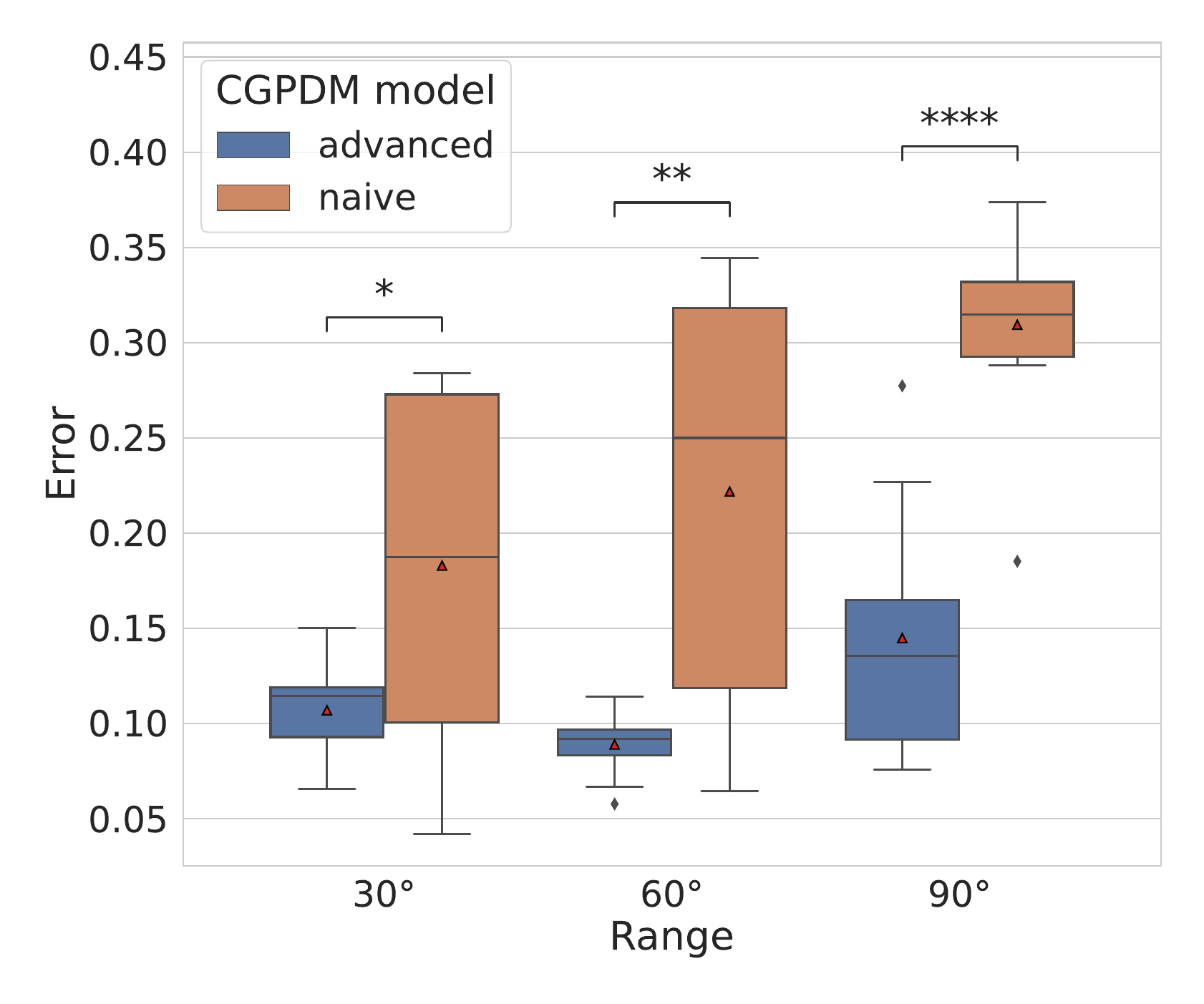}
\caption{Boxplot representing the test prediction errors obtained by the \textit{advanced} and the \textit{naive} CGPDM structures at different oscillation ranges in the real-world experiment. Each configuration was tested following a cross-validation method on a set of 10 cloth trajectories (using 9 for training and one for test). Mean values are indicated with red triangles and statistical significance of T-test comparing \textit{advanced} and \textit{naive} results are represented following the same notation of Fig. \ref{fig:boxplot}}
\label{fig:boxplot_real}
\end{figure}

The models were used to predict the cloth movements obtained in response to the control actions of each test trajectory, measuring the average distance between the real and the predicted mesh points.
In Fig. \ref{fig:real rollout}, we provide a visual representation of the cloth movements, by representing the true and predicted trajectories of a subset of mesh points, in one of the example test cases.
Please refer to the video\footnoteref{note1} for better visualizing the obtained results.
Similarly to the simulated experiment case, Fig. \ref{fig:boxplot_real} represents the observed errors by means of boxplots and the last row of Table \ref{tab:errors} reports the mean distances between true and predicted mesh points obtained in all the considered movement ranges.

\section{Discussion}\label{sec:discussion}
The experimental results obtained in simulation confirm the capacity of CGPDM to capture the cloth dynamics of oscillations along axes Y and Z.
When trained with a sufficient amount of data, CGPDMs obtained satisfying results in a variety of movement ranges.
Training with only 5 sequences seems insufficient to properly capture the considered dynamics.
When training from 10 to 15 sequences, the observed errors diminish significantly; instead working with 20 training trajectories generate minor signs of over-fitting.

For smaller movement ranges ($R=30$° or $R=60$°), the reconstructed trajectories of the mesh of points appear similar to the true ones.
Conversely, for wider ranges ($R=90$° or $R=120$°), discrepancies between true and predicted points begin to be more evident, but the CGPDMs are still able to capture the overall movement of the cloth.

Moreover, the proposed \textit{advanced} CGPDM structure significantly improves accuracy and consistency of the results in the majority of cases, when compared to the \textit{naive} model.
This effect is clearer in a low-data regime and when dealing with wide oscillation ranges.

Finally, results obtained in the real-world experiments confirm the trends observed in the simulated scenario.
The \textit{advanced} CGPDM structure drastically outperforms the \textit{naive} model that seems unable to cope with the high noise that afflicts the real experimental setup.

\section{Conclusion}\label{sec:conclusion}
We presented CGPDM, a modeling framework for high-dimensional dynamics governed by control actions.
Essentially, this model projects observations into a latent space of low dimension, where dynamical relations are easier to infer.
CGPDMs were applied to a robotic cloth manipulation task, where the observations are the coordinates of the cloth mesh.
We tested CGPDMs in both simulated and real experiments.
The observed results empirically demonstrate that the proposed \textit{advanced} CGPDM structure can capture the complex high-dimensional cloth dynamics given a small number of trajectories to learn from by leveraging the data efficiency that characterizes GP-based methods.

In future works, we aim to apply CGPDM within Model-Based Reinforcement Learning algorithms (such as \cite{blackDrops, mcpilco}) to automatically learn control policies for high-dimensional systems. Moreover, CGPDM formulation could be extended through the introduction of back constraints \cite{lawrence2006local} to preserve local distances and obtain an explicit formulation of the mapping from the observation to latent space. 
Finally, the integration of context variables within the CGPDM formulation could permit generalizing over different types of cloth fabric.

\bibliography{references}

\end{document}